\documentclass[runningheads]{llncs}
\usepackage[T1]{fontenc}
\usepackage{hyperref}
\usepackage{graphicx}
\usepackage{bbding}
\usepackage{amsmath}
\usepackage{amsfonts}
\usepackage{amssymb}
\usepackage{cleveref}
\usepackage[linesnumbered,ruled,vlined]{algorithm2e}
\usepackage{tikz}
\usetikzlibrary{arrows.meta,calc,backgrounds,matrix}

\definecolor{nodeblack}{HTML}{111111}
\definecolor{alerttop}{HTML}{8F3634}
\definecolor{alertbottom}{HTML}{D67B76}
\definecolor{alertdraw}{HTML}{BB5651}
\definecolor{alertline}{HTML}{E5908D}

\usepackage{xcolor}
\definecolor{r1}{HTML}{0072B2}    
\definecolor{r2}{HTML}{E69F00}   
\definecolor{r3}{HTML}{CC79A7}   
\definecolor{all}{HTML}{009E73}     

\begin{document}
\title{All Eyes on the Workflow: Automated and Efficient Event Discovery from Video Streams}
\titlerunning{Automated and Efficient Event Discovery from Video Streams}
%
\author{Marco Pegoraro$\,$\Envelope\textsuperscript{,*,}\inst{1}\orcidID{0000-0002-8997-7517} \and
Jonas Seng\textsuperscript{*,}\inst{2}\orcidID{0009-0006-1772-5296} \and
Dustin Heller\inst{2} \and
Wil M.P. van der Aalst\inst{1}\orcidID{0000-0002-0955-6940} \and
Kristian Kersting\inst{2}\orcidID{0000-0002-2873-9152}}
\authorrunning{M. Pegoraro et al.}
%
\institute{Chair of Process and Data Science, RWTH Aachen University,\\Ahornstr. 55, 52074 Aachen, Germany\\
\email{\{pegoraro,wvdaalst\}@pads.rwth-aachen.de} \and
Artificial Intelligence \& Machine Learning Lab, Technical University of Darmstadt,\\Hochschulstr. 1, 64289 Darmstadt, Germany\\
\email{jonas.seng@tu-darmstadt.de, kersting@cs.tu-darmstadt.de}}
\maketitle              
\begin{abstract}
Disciplines such as business process management and process mining aid organizations by discovering insights about processes on the basis of recorded event data. However, an obstacle to process analysis is data multi-modality: for instance, data in video form are not directly interpretable as events. Existing approaches rely on a dictionary of activity label as input, cannot provide frame-by-frame labeling explanations, or rely on superseded computer vision techniques. In this work, we present SnapLog, an approach to extract event data from videos by converting frames to feature vectors using image embeddings and performing temporal segmentation through frame-wise similarity matrices. A generalized few-shot classification is then used to assign labels to the video segments, yielding labeled, timestamped sub-sequences of frames that are interpretable as events. Conventional process mining techniques can be used to analyze the resulting data. We show that our approach produces logs that accurately reflect the process in the videos.         

{\let\thefootnote\relax\footnote{{\textsuperscript{*} These authors contributed equally to this work.}}}
\end{abstract}

\section{Introduction}\label{sec:introduction}

Process mining has become a key instrument for understanding how processes are actually executed in practice. By analyzing event logs, organizations can discover process models, detect deviations, and identify opportunities for improvement. The effectiveness of these analyses, however, depends on the availability of event data in a structured log format. In many real-world settings, this assumption is increasingly violated: process-relevant evidence is collected in multimodal form, such as videos, images, sensor streams, and audio, rather than directly as timestamped event records. For example, in an airport, process-relevant evidence may come from surveillance videos of passenger movements, barcode scans of boarding passes and baggage tags, audio announcements, and sensor data from security checkpoints or baggage handling systems. Rather than being recorded directly as event logs, these heterogeneous data sources cannot directly describe operations in processes such as check-in, security screening, boarding, or baggage transfer. As a consequence, a large and growing share of operational behavior remains inaccessible to conventional process mining methods.

Video data are an important example of this gap. Cameras are now pervasive in manufacturing, logistics, healthcare, retail, and domestic environments, and they capture rich information about the execution of physical activities. However, from the perspective of process mining, videos are difficult to use directly. They encode process behavior implicitly in high-dimensional spatio-temporal signals, rather than explicitly in event abstractions such as activity labels and timestamps. Bridging this representation mismatch is therefore a central challenge: if videos can be transformed into event logs reliably, then process mining can be applied to domains where event data are currently unavailable.

This challenge is also a practical one. Manually annotating videos into process events is labor-intensive, expensive, and difficult to scale. It requires domain expertise, careful temporal segmentation, and consistent labeling across recordings. At the same time, fully supervised video understanding approaches often require large labeled datasets, which are rarely available for process-specific activities. In process analysis scenarios, one may have only a few labeled examples of each activity, while still needing a meaningful event log for downstream mining. This creates a strong need for methods that can exploit modern computer vision embeddings while remaining effective in low-label regimes.

\begin{figure}[t]
    \centering
    \includegraphics[width=\linewidth]{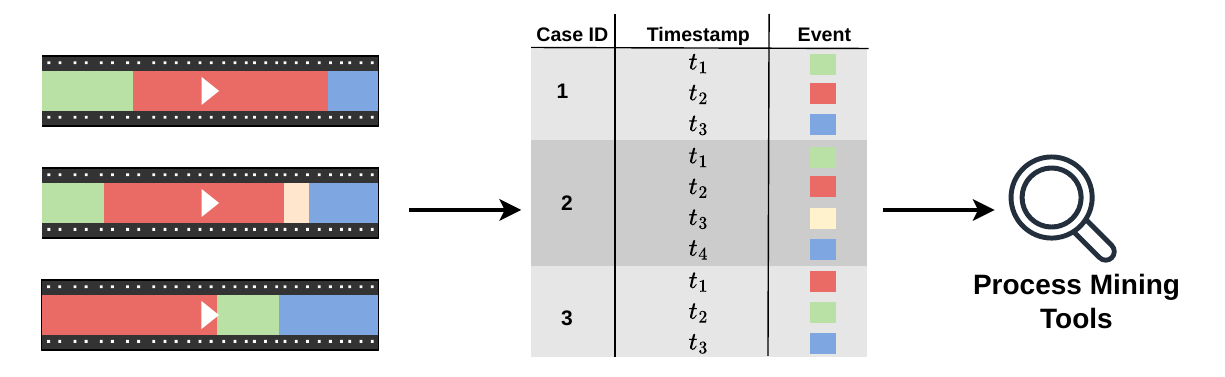}
    \caption{\textbf{SnapLog extracts logs from video data.} Our method, SnapLog, constructs an event log from video data which can then be used downstream by standard process mining algorithms.}
    \label{fig:intro-example}
\end{figure}

In this paper, we address this need by proposing SnapLog, a pipeline that \emph{extracts event logs from video data via temporal segmentation and few-shot classification}. The motivation is twofold. First, we seek to make process mining applicable to video-centric environments by producing standard event-log abstractions from raw visual observations. Second, we aim to do so in a way that is realistic for practice, where labeled examples are scarce and process activities may vary in appearance across recordings. Our approach combines frame-level semantic grouping with generalized few-shot video classification, yielding labeled and timestamped video segments that can be interpreted as process events. The resulting logs can then be analyzed with conventional process mining techniques, enabling process discovery and comparison across multiple recordings (\Cref{fig:intro-example}).

An additional motivation for our approach is that video-based event extraction is inherently \emph{uncertain}; for instance, some clips may support multiple plausible labels. Rather than viewing this uncertainty only as noise, it can be \emph{represented explicitly and passed on to the process mining stage}. In fact, our method can output not only a deterministic event log (via the most likely label per segment), but also an uncertain event log, by retaining the probability distribution over candidate activity labels for each extracted event. This is highly relevant to recent work on uncertain and probabilistic event data, which shows that uncertainty-aware process mining can preserve ambiguity instead of forcing premature decisions~\cite{DBLP:conf/caise/000122}. Our pipeline complements these efforts by providing a computer-vision-based mechanism that can generate such uncertainty-aware logs from raw video observations. The resulting uncertain logs may then be analyzed through specialized analysis techniques available in literature, which are able to account for the probabilistic or non-deterministic information associated to event attributes (in our setting, activity labels).

\textbf{Related work.} Within process mining, video data have first been discussed precisely in the context of uncertain event data by Gal and Cohen, as a natural source for uncertain logs~\cite{DBLP:conf/bpm/CohenG21}.
More generally, the processing of unstructured and multimodal event data is regarded as an important contemporary challenge in the automatic analysis of processes~\cite{DBLP:conf/modellierung/KoschmiderAFILA24}.
Though research on event data abstraction from video data is in its infancy, previous attempts have been made: for instance, Lepsien et al.~\cite{DBLP:conf/bpm/LepsienKK23,agriculture13081639} described a pipeline for extracting events from video feeds in the context of pig farming. While the setting of their work include a focus on object tracking (since they abstract a process case with a single pig) and a fully known activity label dictionary, in our work we consider separate videos depicting process cases (thus, with a focus on segmentation) and a partially known dictionary of labels (motivating the few-shot classification). Other existing approaches adopt substantially different designs, including direct frame-agnostic video-to-activities extraction~\cite{DBLP:conf/caise/CorradiniP0RS25} or a general software architecture for log extraction from videos~\cite{DBLP:journals/dss/KratschKR22,DBLP:journals/dss/WordehoffEKKR25}, rendering both methodologically highly distinct from our work. A recent approach to video-based process mining~\cite{DBLP:conf/wecwis/GavricBP24} relies on a heavy cascade of Visual Language Models (VLMs) and LLMs. While this enables zero-shot classification, the high inference costs and dependency on external API providers create significant barriers for large-scale industrial use. Furthermore, such pipelines often necessitate massive cloud-based compute clusters, complicating the handling of sensitive organizational data. In contrast, SnapLog is designed for resource-constrained, on-premise deployment. By utilizing lean models optimized for few-shot adaptation ($<20$ samples), we eliminate the need for specialized high-performance computing (HPC) infrastructure. Our approach can be executed on standard local hardware with one or few GPUs, offering a cost-effective and privacy-preserving alternative that is orders of magnitude more scalable for extensive video archives. SnapLog and the logs resulting from the evaluation are openly available online\footnote{\url{https://github.com/J0nasSeng/SnapLog}}.

The remainder of the paper is structured as follows. \Cref{sec:preliminaries} introduces the basic computer vision and process mining notions needed to describe our approach. \Cref{sec:method} illustrates our video-to-log pipeline. \Cref{sec:evaluation} presents the results of a quantitative evaluation of our extraction method. Lastly, \Cref{sec:conclusion} concludes the paper.

\section{Preliminaries}\label{sec:preliminaries}

The video-to-log pipeline requires a combination of concepts from the fields of machine learning, computer vision, and process mining. In this section, we present the necessary preliminary notions.

\textbf{Event Data}. Our target representation is the event log, a collection of records tracking the activities that take place in a process. In this work, we will only consider the control-flow, i.e., each event is abstracted through an activity label, a timestamp marking the beginning of the event, and the timestamp at completion. Formally, for a set of activity labels $\mathcal{A}$ and a totally ordered universe of timestamps $\mathcal{T}$. Then, an event is a triple $e = (a, t_\mathit{start}, t_\mathit{end}) \in (\mathcal{A} \times \mathcal{T} \times \mathcal{T})$, a trace $\sigma \in \mathcal{P}(\mathcal{A} \times \mathcal{T} \times \mathcal{T})$ is a set of events, and an event log $L \in \mathcal{B}(\mathcal{P}(\mathcal{A} \times \mathcal{T} \times \mathcal{T}))$ is a multiset of traces. Conversely, in uncertain events activity labels are probabilistic: $e_U = (p_{e_U}, t_\mathit{start}, t_\mathit{end})$, where $p_{e_U}(a)$ is a discrete probability distribution over the set of labels $\mathcal{A}$. An uncertain trace $\sigma_U$ is a set of uncertain events, and an uncertain log $L_U$ is a multiset of uncertain traces.

Our starting point for event log extraction will be a set of videos, for which we will present a processing pipeline.

\textbf{Video Data and Feature Embeddings.} Consider an input video $\mathbf{X} \in \mathbb{R}^{C \times T \times H \times W}$ consisting of $T$ frames with height $H$, width $W$, and $C$ color channels. The common practice to solve downstream tasks such as classification on video data is to first encode the input video into a dense, lower-dimensional vector representation, a so-called embedding. This representation efficiently encodes the spatio-temporal information required to solve the downstream task. Typically, neural networks are used as encoders; specifically, convolutional neural networks (CNNs) and Vision Transformers (ViT) have been shown to yield embeddings useful for many downstream tasks~\cite{DBLP:journals/csur/SchiappaRS23,DBLP:journals/corr/abs-2405-03770}.

Since our pipeline for log extraction consists of first splitting a video into semantically meaningful subsequences, followed by classifying these subsequences to obtain event labels, we employ two encoding mechanisms: One that fully preserves temporal structure (for semantic grouping of frames) and one for final classification (event label assignment).

To split the video into semantically meaningful subsequences, we rely on clustering frame-wise representation. Since the Vision Transformer (ViT) architecture has been shown to learn robust, state-of-the-art features from image data that produce well-separated and semantically grounded clusters in an Euclidean-like embedding space~\cite{DBLP:conf/nips/RaghuUKZD21,DBLP:journals/csur/SchiappaRS23}, we leverage a pre-trained ViT~\cite{DBLP:conf/iclr/DosovitskiyB0WZ21} in our pipeline. Given a video $\mathbf{X} \in \mathbb{R}^{T \times H \times W \times C}$ consisting of $T$ discrete frames, each frame $\mathbf{x}_t$ is processed independently to generate a frame-level embedding. Formally, an input image $\mathbf{x} \in \mathbb{R}^{H \times W \times C}$ is reshaped into a sequence of $N$ flattened 2D patches $\mathbf{x}_p \in \mathbb{R}^{N \times (P^2 \cdot C)}$, where $P$ is the patch size and $N = \frac{HW}{P^2}$. These patches are mapped to a $D$-dimensional latent space via a linear projection $\mathbf{W}_e \in \mathbb{R}^{(P^2 \cdot C) \times D}$. To maintain spatial context, a position embedding $\mathbf{e}_{pos}$ is added to the projected patches. The resulting sequence, augmented by a learnable $\texttt{[CLS]}$ token, serves as the input to a Transformer encoder:$$\mathbf{Z}_{in} = [\mathbf{x}_{class}; \mathbf{x}_p^1\mathbf{W}_e; \mathbf{x}_p^2\mathbf{W}_e; \dots; \mathbf{x}_p^N\mathbf{W}_e] + \mathbf{e}_{pos}$$The encoder applies $L$ layers of Multi-Head Attention (MHA) and Multi-Layer Perceptrons (MLP). In this work, we utilize the state of the $\texttt{[CLS]}$ token at the final layer $L$ as the representative embedding for the frame~\cite{DBLP:conf/iclr/DosovitskiyB0WZ21}.

To obtain final event classifications, the R(2+1)D architecture is used, which encodes video data while preserving crucial spatio-temporal information, thereby enabling high classification performance~\cite{DBLP:conf/cvpr/TranWTRLP18}. To facilitate downstream log extraction, the video is encoded by a function $\mathbf{Z} = f(\mathbf{X})$, where $\mathbf{Z} = (\mathbf{z}_1, \mathbf{z}_2, \dots, \mathbf{z}_T)$ represents a sequence of $T$ spatiotemporal embeddings. The function $f$ is realized through the R(2+1)D architecture, which factorizes a standard 3D convolutional filter of size $d \times k \times k$ into two separate operations. Specifically, for each convolutional block, the architecture first applies a spatial convolution with a weight tensor of size $1 \times k \times k$, immediately followed by a temporal convolution with a weight tensor of size $d \times 1 \times 1$. Formally, if $\ast_{s}$ denotes the spatial convolution and $\ast_{t}$ denotes the temporal convolution, the intermediate transformation for a hidden layer $h$ is given by:$$\mathbf{h}_{i+1} = \sigma(\ast_{t} (\sigma(\ast_{s} (\mathbf{h}_i))))$$where $\sigma$ denotes a non-linear activation function, such as ReLU. This specific factorization better preserves spatiotemporal information than full 3D convolutions. A final global temporal pooling layer is used to obtain a fixed-sized representation $\mathbf{z} \in \mathbb{R}^d$ that can be used for downstream few-shot classification~\cite{DBLP:conf/cvpr/TranWTRLP18}.

\textbf{Temporal Action Segmentation (TAS).} The objective of TAS is to partition an untrimmed video into a sequence of semantically cohesive segments, representing distinct process activities or states~\cite{DBLP:journals/pami/DingSY24,DBLP:journals/ijwmip/WangZQ24,DBLP:journals/pami/WangZYLL24}. Given the sequence of embeddings $\mathbf{Z} = (\mathbf{z}_1, \mathbf{z}_2, \dots, \mathbf{z}_T)$ extracted in the previous steps, we employ an unsupervised approach to cluster frames that exhibit high semantic similarity. We model the TAS task as a clustering problem. Because the embeddings $\mathbf{Z}$ are obtained from a ViT encoder whose embedding space exhibits a geometry similar to the Euclidean space~\cite{DBLP:conf/nips/RaghuUKZD21,DBLP:journals/csur/SchiappaRS23}, we opt for the K-Means algorithm as a clustering algorithm. K-Means has been proven to work well in Euclidean spaces with hundreds of dimensions~\cite{DBLP:conf/iccv/NamAKHC21}, while e.g., density-based clustering tends to struggle in such high-dimensional spaces~\cite{DBLP:journals/corr/abs-2601-08841}. Formally, for a set of embeddings $\mathbf{Z}$, K-Means seeks to partition the observations into $k$ sets $S = \{S_1, S_2, \dots, S_k\}$ so as to minimize the within-cluster sum of squares (WCSS):$$\arg\min_{S} \sum_{i=1}^{k} \sum_{\mathbf{z} \in S_i} \| \mathbf{z} - \mu_i \|^2$$where $\mu_i$ is the centroid of points in $S_i$. This approach provides a low-complexity method for splitting a given set of frame embeddings into semantically meaningful subsequences, enabling downstream event classification. 

\textbf{Few-Shot Video Classification.} Few-shot video classification is a method for an existing classification model to learn new unseen classes using only a few new examples~\cite{DBLP:journals/csur/FerdausNTAI26,DBLP:journals/ijcv/WanyanYDX25}. In our pipeline, we adopt a few-shot video classification technique inspired by the generalized approach illustrated in~\cite{DBLP:journals/pami/XianKDTSA22}. In this approach, it is assumed that there is a small dataset of labeled videos. These videos show examples of new classes the model should learn to recognize. The pre-trained backbone is used to obtain a video encoding $\mathbf{z}$ which is then fed into a linear classifier $\mathbf{W}_c$. The classifier $\mathbf{W}_c$ is then learned (or fine-tuned) by minimizing cross-entropy loss on the dataset showing the new classes. Therefore, adding new classes to the model to be recognized reduces to learning a small linear model that leverages the encodings of the previously pre-trained encoder.

\section{Method}\label{sec:method}

\begin{figure}[t]
    \centering
    \includegraphics[width=\linewidth]{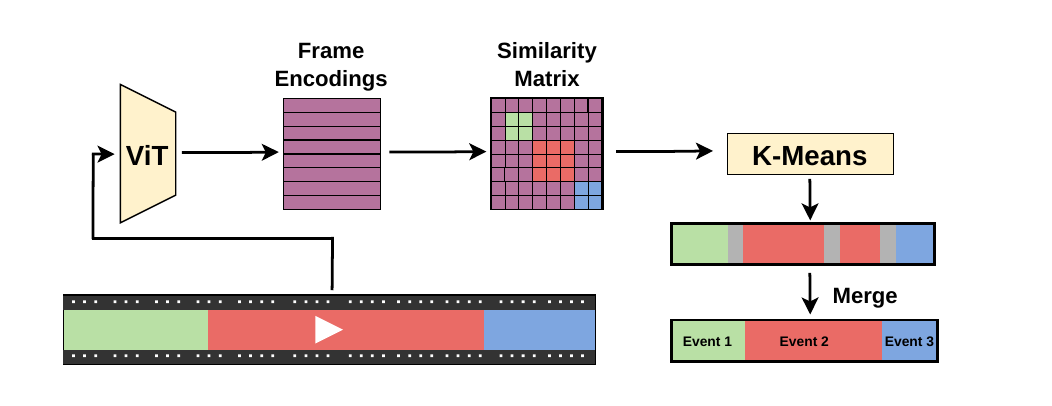}
    \caption{\textbf{Video Segmentation.} From left to right: After the extraction of frame-wise embeddings using a ViT, the embeddings are used to compute a similarity matrix. By applying K-Means on the similarity matrix, atomic events are formed. These are merged into larger action segments using a greedy merging algorithm that pairs up atomic events based on their proximity in the embedding space, thus achieving a semantically grounded grouping.}
    \label{fig:segmentation}
\end{figure}

\begin{figure}[t]
    \centering
    \includegraphics[width=\linewidth]{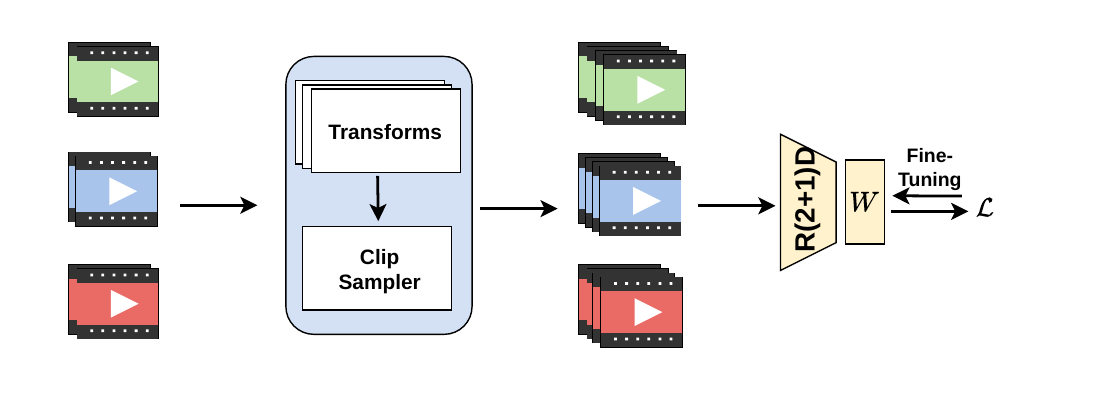}
    \caption{\textbf{Finetuning.} Given a few labeled examples for novel classes, we first perform data augmentation by applying transformations (gray-scale, rotations, etc.) to increase the number of given labeled examples. Then, a new head is added to the R(2+1)D encoder and learned via gradient descent.}
    \label{fig:finetuning}
\end{figure}

Our video-to-log pipeline consists of two key stages: the temporal action segmentation (\Cref{fig:segmentation}), and the few-shot video classification (\Cref{fig:finetuning}). We now describe the different stages of our pipeline in detail.

\subsection{Temporal Action Segmentation}

In our pipeline, Temporal Action Segmentation (TAS) aims to identify subsequences in a given video that represent consecutive frames of events in an unsupervised way. Treating TAS as an unsupervised learning problem enables us to detect potentially relevant events without the expense and time required for labeling. To perform TAS, our pipeline first applies a pre-trained ViT to each video frame to obtain a sequence of frame embeddings that compactly represent the content of each frame. These embeddings are then clustered using K-means, yielding semantically meaningful subsequences of the input video via unsupervised clustering. These subsequences are then passed downstream for classification.

To find semantically meaningful subsequences in $\mathbf{X}$, we aim to identify event boundaries, e.g., when frames changes occur. Inspired by the approach in~\cite{DBLP:conf/iccv/NamAKHC21}, K-Means clustering is used to find atomic events by clustering $\mathbf{Z}$. A simple merge algorithm combines extremely short events, reducing fragmentation in the action segmentation.

\textbf{Embedding Extraction.} Our approach employs a frame-wise feature encoding of an input video. A pre-trained ViT is used to compute $\mathbf{Z} = (\mathbf{z}_1, \dots, \mathbf{z}_T)$ where each $\mathbf{z}_i = f(\mathbf{x}_i)$ is the frame embedding of the $i$-th image computed by the ViT $f$. Thus, each frame in a video $\mathbf{X} \in \mathbb{R}^{T \times H \times W \times C}$, represented as a sequence of RGB images $(\mathbf{x}_1, \dots, \mathbf{x}_T)$, is encoded independently. Here, $T$ is the number of frames in the video, $H$ is the frame height, $W$ is the frame width, and $C$ is the number of channels (3 for RGB images). The embedding sequence $\mathbf{Z}$ is a real-valued matrix with dimension $T \times d$ where $d$ is the dimension of the latent space learned by the ViT. The ViT was pre-trained on Imagenet, a large-scale corpus designed for image classification, often used to pre-train vision models.

Encoding the video frame-by-frame was a deliberate choice as it enables us to assign each frame to a specific cluster when identifying semantically meaningful subsequences. Since the event classification requires a small labeled dataset of subsequences showing specific events, we can directly show identified subsequences to a human user who can provide the required labels and assess the quality of identified subsequences.

\textbf{The Frame-Wise Similarity Matrix.} After computing frame-wise embeddings, semantically similar frames must be grouped into subsequences of the original video to identify events. To achieve this, a similarity matrix $\mathbf{D} \in \mathbb{R}^{T \times T}$ is computed. This matrix contains pairwise distances between the feature vectors for all frames of a video. 
The cosine similarity measure is used to compute each element of $\mathbf{D}$, i.e., $d_{i j} =  1 - \frac{\mathbf{z}_i \cdot \mathbf{z}_j}{||\mathbf{z}_i||||\mathbf{z}_j||}$.
The clustering algorithm must consider the temporal axis since frames closer together w.r.t. the time dimension should be more likely clustered together. This allows repeated actions to be clustered separately as the clustering algorithm will separate them as two events with the same action. To achieve this, the frame index of each frame is appended as a new row of $\mathbf{D}$. Thus, we obtain a matrix $\mathbf{D}' = \mathbf{D} || \mathbf{i}$ where $||$ denotes the row-wise concatenation and $\mathbf{i} = (1, \dots, T)$.
Next, $\mathbf{D}'$ is normalized as $d_{i j}' = \frac{d_{i j}'}{\mathbf{D}'_i}$, and a softmax operation is performed, i.e.,  $d_{i j}' = \frac{\text{exp}(d_{i j}')}{\sum_k \text{exp}(d_{i k}')}$.

Note that each column of $\mathbf{D}'$ is a contextualized representation of a frame w.r.t. the entire video: Each column $j$ represents one frame and encodes distances of $j$ to all other frames, thereby contextualizing $j$. While the sequence of encoded frames $\mathbf{Z}$ could also be used to cluster frames, $\mathbf{Z}$ does not provide a contextualized representation of each frame. Adding global context has been shown to stabilize clustering~\cite{DBLP:conf/iccv/NamAKHC21}. 

\textbf{Temporal Event Clustering.} To cluster temporal events from the similarity matrix $\mathbf{D}'$, K-Means is applied, which assigns a cluster index $c \in \{1, \dots, k\}$ to each frame. The clusters found by K-Means yield \textit{atomic} events, i.e., consecutive frames with the same cluster assignment. We term these events atomic, as K-Means tends to find short events spanning too few frames. However, simply setting $k$ to a larger value, results in too coarse clusters, often spanning frames of multiple events. Since $k$ is the only parameter we can control in K-Means, it does not provide fine-grained control over the length of clustered events. Hence, a simple yet effective merging algorithm is used to merge shorter atomic events into adjacent events to form composite events. These events then form the subsequences that are ultimately classified in the downstream few-shot classification.



Merging atomic events is done by first defining a process-dependent minimum event length $l_{min}$. We then merge events that are shorter than $l_{min}$ into a larger event using a greedy approach: For each sub-threshold event, the algorithm identifies its chronologically adjacent neighbors.
If an event has only one neighbor (i.e., is at the start or end of the video), it is merged with its neighboring event. If an event is flanked by two segments, it is merged with the neighbor whose cluster centroid exhibits the smallest Euclidean distance to its own. This ensures that our merging strategy combines atomic events that semantically belong together with high probability. In cases where neighbors differ in their cluster assignment and the distances are equidistant, the algorithm prioritizes the preceding neighbor to maintain temporal consistency. If both neighbors share the same cluster assignment, the event inherits that assignment, again striving to maximize probability of combining atomic events that semantically belong together. This iterative pass ensures that the final sequence consists only of events that satisfy the minimum length requirement. Note that each step of our greedy merging strategy is computationally efficient while it aims to pair up atomic events that are too short to be plausible alone but semantically belong to the same event of the process. See Algorithm~\ref{algo:event_merging} for the pseudocode.

\begin{algorithm}[t]
\caption{Temporal Event Merging and Refinement}\label{algo:event_merging}
\SetKwInOut{Input}{Input}\SetKwInOut{Output}{Output}
\Input{Sequence of atomic events $E$, threshold $l_{min}$, cluster centroids $C$}
\Output{Refined sequence of composite events $E'$}

\ForEach{event $e_i$ in $E$}{
    \If{length of $e_i < l_{min}$}{
        Identify neighbors $N$ adjacent to $e_i$ on the temporal axis\;
        \eIf{$|N| == 1$}{
            Merge $e_i$ into the single neighbor\;
        }{
            $n_{prev}, n_{next} \gets$ neighbors in $N$\;
            \eIf{cluster assignment of $n_{prev} ==$ cluster assignment of $n_{next}$}{
                Merge $\{n_{prev}, e_i, n_{next}\}$ into one composite event\;
            }{
                $d_{prev} \gets$ distance between centroids of $e_i$ and $n_{prev}$\;
                $d_{next} \gets$ distance between centroids of $e_i$ and $n_{next}$\;
                \eIf{$d_{prev} \leq d_{next}$}{
                    Merge $e_i$ into $n_{prev}$\;
                }{
                    Merge $e_i$ into $n_{next}$\;
                }
            }
        }
    }
}
\Return $E'$\;
\end{algorithm}

\subsection{Few-Shot Video Classification}
Once our pipeline has segmented the video into subsequences potentially containing frames representing events using TAS, we apply a few-shot event classification next to assign an event label to each identified segment.

\textbf{Event Classification} To classify the identified video segments, we employ a flexible few-shot approach that facilitates rapid adaptation to new event types, akin to~\cite{DBLP:journals/pami/XianKDTSA22}. This rapid adaptation allows our approach to perform well with only a few labeled instances, enabling cheap and fast adaptation of pre-trained models. We utilize a 34-layer R(2+1)D encoder $g$, pre-trained on the Sports-1M dataset~\cite{DBLP:conf/cvpr/KarpathyTSLSF14}, to map an event $\mathbf{X_e} \in \mathbb{R}^{T_e \times H \times W \times C}$ from the output of Alg.~\ref{algo:event_merging} (i.e., $\mathbf{X_e} \in E'$) to a dense latent representation $\mathbf{z} \in \mathbb{R}^d$, where $\mathbf{z} = g(\mathbf{X})$. A linear classification head $\mathbf{W}_f \in \mathbb{R}^{d \times m}$ then produces the class distribution via $p(a | \mathbf{X_e}) = \text{softmax}(\mathbf{W}_f \cdot \mathbf{z})$ for classes $a \in \mathcal{A}$, which we interpret as activities.

\textbf{Model Adaptation and Augmentation.} To optimize the encoder for domain-specific process mining, we employ a two-stage adaptation strategy that separates feature refinement from task specialization. Initially, we fine-tune the encoder $g$ using a base dataset $\mathcal{D}_b$. This stage incentivizes $g$ to extract spatio-temporal embeddings that are semantically aligned with the target environment, effectively bridging the gap between general-purpose video features and the nuances of process-related actions. Following this refinement, the model undergoes few-shot adaptation to the specific task defined by a limited labeled dataset $\mathcal{D}_c$. During this phase, the encoder $g$ is frozen to preserve the learned domain features, while only the new classification head $\mathbf{W}_f$ is optimized via cross-entropy loss. This separation of concerns prevents the model from overfitting to the sparse samples in $\mathcal{D}_c$ while allowing it to learn new activity classes rapidly. To further bolster model robustness against the limited sample size, we apply consistent stochastic augmentations across all frames. These include geometric rotations $r \in \{90^\circ, 180^\circ, 270^\circ\}$, photometric brightness scaling $b \in [0.5, 1.5]$, Gaussian noise injection, and grayscale conversion. This augmentation pipeline ensures that the learned process embeddings remain invariant to the environmental and sensor-induced variability common in real-world video streams.

\textbf{Clip Sampling Strategy.} Training and evaluation are performed on clips of 16 consecutive frames extracted from the event sequences. We chose 16 frames to capture enough informative frames for accurate event classification while keeping the computational load as small as possible. We compare two distinct sampling regimes: non-overlapping and overlapping clips. While non-overlapping sampling ensures each frame belongs to a unique clip, the overlapping strategy allows frames to reappear across non-identical clips. This latter approach significantly increases the volume and diversity of training data available per epoch, which is particularly beneficial in the few-shot setting.

\textbf{Event Log Abstraction}. Lastly, we abstract the labeled segments as events. We build an event log $L$ by creating a trace $\sigma \in L$ for each video $\mathbf{X}$. Let $E'$ be the list of segments corresponding to the video $\mathbf{X}$. For the segment $\mathbf{X_e}$, let $t^{\mathbf{X_e}}_\mathit{start}$ (respectively, $t^{\mathbf{X_e}}_\mathit{end}$) be the timestamp of the first (respectively, last) frame of $\mathbf{X_e}$. Then, we may abstract a traditional (certain) process trace through maximum likelihood estimation, with $\sigma = \{(\arg\max_{a} p(a | \mathbf{X_e}), t^{\mathbf{X_e}}_\mathit{start}, t^{\mathbf{X_e}}_\mathit{end}) \mid \mathbf{X_e} \in E'\}$. Since we know the probability distribution of all the labels, we have the option of recording it in the dataset, obtaining an uncertain event log with probabilistically uncertain activity labels: $\sigma_U = \{(p(a | \mathbf{X_e}), t^{\mathbf{X_e}}_\mathit{start}, t^{\mathbf{X_e}}_\mathit{end}) \mid \mathbf{X_e} \in E'\}$. Note that, in both the certain and uncertain event log translations, segments have unique start and end timestamps, and the resulting trace will have one distinct event per segment. This ensures that no pair of events in a trace are associated to overlapping timestamp intervals; as a consequence, extracted timestamps are certain and result in totally ordered traces. This results in an uncertain log of type $[A]_\mathbb{W}$ in the taxonomy of~\cite{DBLP:journals/is/PegoraroUA21}.

\section{Evaluation}\label{sec:evaluation}
The goal of our evaluation is to quantify the fidelity of the event logs extracted from raw video and their utility for downstream process mining. We adopt a multi-stage validation strategy to isolate the performance of each pipeline component: first, we assess TAS accuracy; second, we evaluate action classification on the resulting segments. Finally, we demonstrate that the synthesized event logs successfully reconstruct a meaningful and representative process model, proving the end-to-end viability of our lean, few-shot approach for real-world video data.

\subsection{Experimental Setup}
\textbf{Datasets.} We evaluate our pipeline on unconstrained real-world environments using two datasets.

The TUM Kitchen Dataset \cite{DBLP:conf/iccvw/TenorthBB09} serves as our baseline for high-fidelity process extraction in a stable environment. We selected 16 videos of a brownie-baking process, characterized by a consistent background and fine-grained annotations. To bridge the gap between low-level atomic actions (e.g., ``take spoon'') and the higher-level activities required for meaningful process mining, we utilized a standardized relabeling step. By leveraging an LLM to aggregate these segments into 16 compound activities (e.g., ``prepare batter''), we transformed the raw action recognition data into a structured format suitable for process discovery. The LLM was used to generate aggregation rules \textit{once}, grounded in label semantics. These rules were then manually checked for correctness and applied to the entire dataset to ensure consistency of the new labels.

The Epic Kitchens-100 Dataset \cite{DBLP:journals/ijcv/DamenDFFKMMMPPW22} provides a rigorous test of our pipeline’s robustness to environmental variability. Unlike the TUM dataset, Epic Kitchens features ``in-the-wild'' recordings across 45 diverse kitchen setups. We sampled 24 videos focusing on ``kitchen cleaning'' across 10 different environments. While most egocentric datasets suffer from extreme task variance (e.g., different recipes), cleaning behaviors exhibit the structural consistency necessary for reliable process modeling. This selection allows us to demonstrate that our few-shot approach can generalize across shifting backgrounds and heterogeneous camera angles without requiring extensive retraining for each new scene.

\textbf{Model Adaptation.} Our experimental evaluation utilizes 16 videos from the TUM dataset and 24 from Epic Kitchens, yielding 182 and 294 segments respectively. This distribution presents a rigorous low-data challenge, with an average of only 11 and 7 segments per class. To evaluate few-shot performance, we partition these into base, test, and novel splits; the TUM dataset comprises 114, 22, and 46 segments, while Epic Kitchens contains 196, 24, and 74.
To increase robustness of our model and our evaluation, we apply stochastic augmentations—including rotation, brightness scaling, and noise—to enhance model invariance across these sparse samples. This process expands the TUM segments to 570 (base), 110 (test), and 230 (novel). Similarly, the Epic Kitchens augmented splits reach 980 (base), 120 (test), and 370 (novel).

We utilize a 34-layer R(2+1)D backbone pre-trained on Sports1M~\cite{DBLP:conf/cvpr/KarpathyTSLSF14}, processing 16-frame RGB clips at $112 \times 112$ resolution. The architecture is first fine-tuned on base videos (using Adam~\cite{DBLP:journals/corr/KingmaB14}, $\eta=0.001$, 10 epochs) to align the features with the target process domain. To maximize temporal coverage in training, we sample either 10 non-overlapping or 100 overlapping clips per segment.

For few-shot adaptation to novel activities, we optimize a linear classifier ($\eta=0.01$, 10 epochs) on novel videos while keeping the backbone frozen. At inference, we aggregate predictions across multiple clips per segment to ensure a stable event classification, reporting Top-1 and Top-3 accuracies as indicators of the log's reliability for downstream process discovery. For a given $k$, Top-$k$ accuracy measures the percentage of instances where the true label is among the $k$ highest-probability predictions generated by the model.


\subsection{Results}
\begin{figure}[t]
    \centering
    \includegraphics[width=\linewidth]{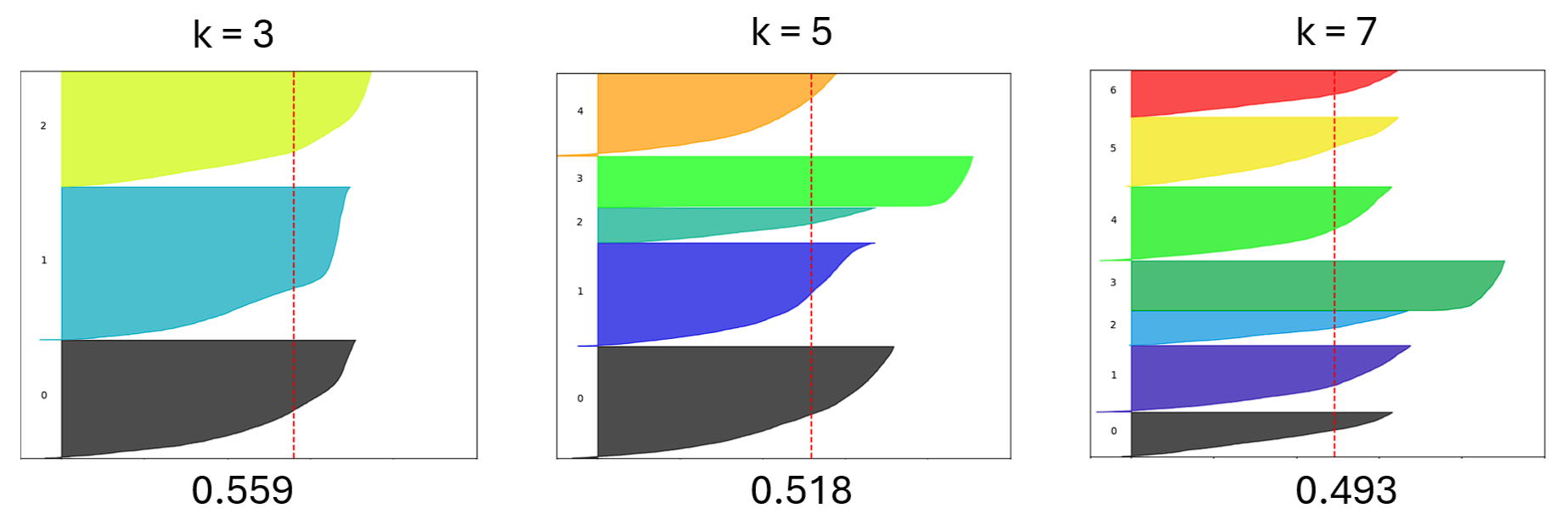}
    \caption{\textbf{Silhouette scores results.} The average silhouette scores of a single video from the TUM dataset for \(k = 3\), 5, and 7 clusters (average represented by the red line). The thickness of the cluster bars indicates the cluster size and the length of the cluster indicates a larger separation to other clusters. $k = 7$, although having a lower average silhouette score, provides the most balanced clustering for our purposes.}
    \label{fig:silhouette-plots}
\end{figure}

\textbf{Segmentation.} Our analysis of the temporal action segmentation reveals that a cluster count of $k=7$ provides the most balanced representation of the feature space for both datasets. Silhouette score evaluations across $k \in \{3, 5, 7\}$ indicate that while lower values of $k$ yield marginally higher scores ($0.563$ for TUM and $0.544$ for Epic Kitchens at $k=3$), they fail to capture the underlying cluster diversity (see~\Cref{fig:silhouette-plots}).

\begin{figure}[t]
    \centering
    \includegraphics[width=\linewidth]{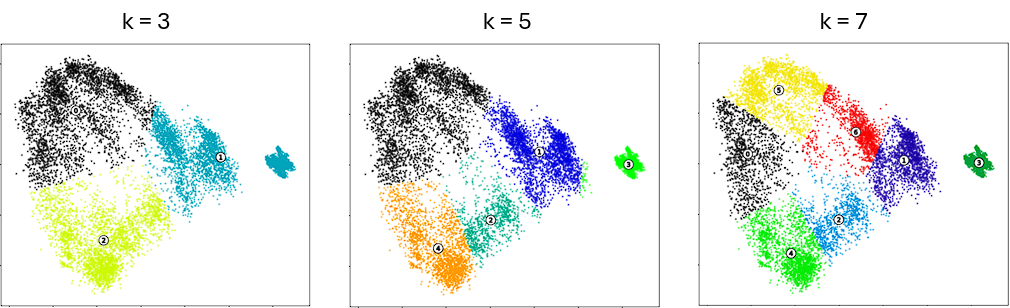}
    \caption{\textbf{Our pipeline produces semantically meaningful video segmentations.} Applying K-Means on encoded frames yields semantically meaningful segmentations. It can be seen that $k = 7$ yields approximately equally sized clusters, indicating a good choice of $k$. Here, we show a projection of the clustered embeddings to a 2-d space using Principal Component Analysis (PCA).}
    \label{fig:k-means-example}
\end{figure}

At $k=7$, silhouette scores for both TUM ($0.498$) and Epic Kitchens ($0.487$) approach the ideal $0.5$ threshold. This configuration results in uniform cluster thickness and superior spatial distribution in the PCA-reduced feature space, preventing the overextension of clusters observed at lower $k$ values (see~\Cref{fig:k-means-example}).

By validating these segments against the ground truth annotations, we find that the pipeline achieves a frame-wise accuracy of $74.3\%$ for TUM and $67.7\%$ for Epic Kitchens. Despite the complexity of the unconstrained environments, the segmentation preserves the structural fidelity of the underlying processes. Most action boundaries align closely with the ground truth, ensuring that the essential sequence of activities remains intact for subsequent process discovery. 

\textbf{Classification Results.} Our few-shot classification results demonstrate that high-fidelity event labels can be extracted even within the extreme constraints of a low-data regime. Despite having only a few videos available for the classification of novel activities, the R(2+1)D backbone—averaged over four random seeds—yielded stable performance with a standard deviation of only $2\%$--$3\%$.

The analysis in Table \ref{tab:few-shot-results} highlights that temporal coverage is a critical performance driver. Transitioning from 10 non-overlapping to 100 overlapping clips per segment improved Top-3 accuracy by up to $23\%$ for the Epic Kitchens dataset. When further bolstered by spatial augmentations, Top-3 accuracies reached $90\%$ for TUM and $85\%$ for the more environmentally complex Epic Kitchens.

These Top-3 metrics are particularly salient given the inherent uncertainty of video-based extraction. As motivated in our introduction, certain segments support multiple plausible interpretations. Rather than forcing a potentially erroneous ``hard'' decision, our pipeline preserves this ambiguity by retaining the probability distribution across the top candidates. This high Top-3 accuracy confirms that our vision-based mechanism reliably captures the true activity within a narrow set of probable labels, providing a robust foundation for uncertainty-aware process mining that preserves semantic richness instead of discarding it as noise. The uncertain logs produced by SnapLog can reflect the Top-3 accuracy, by retaining the three most likely activity labels for each event. Their respective probability may be annotated as-is or, for a more refined result, may be normalized to sum to 1, rendering them immediately interpretable as discrete probability distributions.

\begin{table}[t]
    \centering
    \caption{The averaged results of the few-shot classification. Four methods are evaluated varying between including augmented videos (Aug) or only using base videos (Base) and sampling 10 non-overlapping clips per epoch (10-Clips) or 100 overlapping clips (100-Clips).}
    \begin{tabular}{ccccc}
         \textbf{Accuracy \%} & \multicolumn{2}{c}{\textbf{TUM}} & \multicolumn{2}{c}{\textbf{Epic Kitchens}} \\ \hline
         \textbf{Method} & \textbf{Top-1} & \textbf{Top-3} & \textbf{Top-1} & \textbf{Top-3} \\
         Base-10-Clips & $33.1\pm2.0$ & $63.1\pm3.1$ & $28.2\pm\textbf{2.1}$ & $47.5\pm2.9$ \\
         Base-100-Clips & $50.7\pm\textbf{2.0}$ & $73.7\pm\textbf{2.5}$ & $42.3\pm2.8$ & $70.5\pm\textbf{2.4}$ \\
         Aug-10-Clips & $59.9\pm2.7$ & $82.9\pm2.8$ & $52.2\pm2.1$ & $78.7\pm2.6$ \\
         Aug-100-Clips & $\textbf{67.1}\pm2.7$ & $\textbf{90.1}\pm2.8$ & $\textbf{63.8}\pm2.8$ & $\textbf{85.4}\pm2.9$ \\
    \end{tabular}
    \label{tab:few-shot-results}
\end{table}

\begin{figure}[t]
    \centering
    \includegraphics[width=0.725\linewidth]{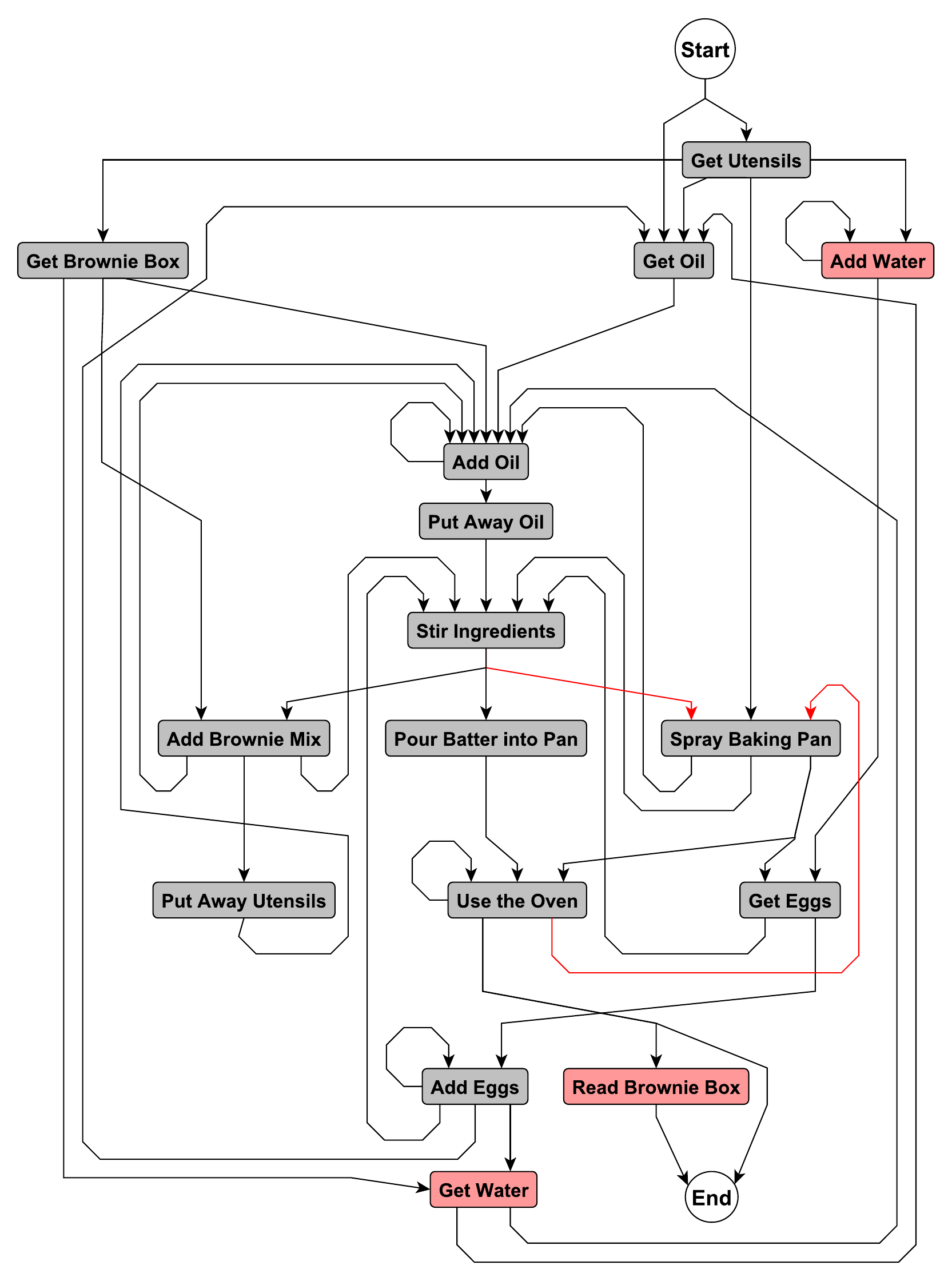}
    \caption{\textbf{A resulting process model.} An example of directly-follows graph extracted from our validation in the non-augmented experiment on the TUM dataset. Red components show faults.}
    \label{fig:tum-example}
\end{figure}

\textbf{Visual Inspection.} While a full quantitative evaluation of the results of process discovery applied to the output log requires a complete case study, and is thus out of the scope of this paper, an example of a simple process model (discovered with the Process Explorer tool) of the log extracted from the TUM dataset can be seen in \Cref{fig:tum-example}.


In general, the model shows a correct procedure to bake brownies, and most activities and connections are sensible. The logical order of cooking steps is well represented in the graph, and the process of baking brownies from a brownie mix is described with good completeness.

Some inconsistencies in ordering do appear, with \textit{Add Water} followed by \textit{Get Water}. The activity \textit{Read Brownie Box} appears near the end, while it should occur near the start. There are some erroneous backward edges to \textit{Spray Baking Pan}. Lastly, note that the activity \textit{Use the Oven}, present in the annotations, labels both the act of preheating the oven and of inserting the pan in the oven, explaining the connection between \textit{Spray Baking Pan} and \textit{Use the Oven.}

In summary, despite some inaccuracies, a clear flow of the process of baking brownies can be identified through the model, which represents the overarching process throughout the videos relatively reliably. 

\section{Conclusion}\label{sec:conclusion}

This paper presented a pipeline for extracting process-relevant event logs from raw video through temporal segmentation and few-shot activity classification. By converting unstructured video observations into structured, timestamped events, we enable the application of process mining in domains where digital footprints are traditionally absent. Our approach is designed to be deployable on standard, non-HPC infrastructure, requiring only a small labeling effort—approximately 15–20 videos—to adapt to new environments.

A key contribution of this work is the generation of uncertainty-aware event logs. By retaining class probability distributions, we preserve semantic ambiguity for downstream analysis rather than forcing premature, deterministic decisions. This is supported by our high Top-3 classification accuracies, which ensure that the true activity is captured within a narrow set of candidates even in complex kitchen environments.

\textbf{Limitations and Future Work.}
Current limitations include the necessity for initial human labeling and a modular design that is not yet end-to-end differentiable. While our labeling requirements are low, future iterations could leverage VLMs in-the-loop to further automate the initial annotation process while maintaining the efficiency of a lean inference pipeline. Developing a fully differentiable framework would also allow for joint optimization of the segmentation and classification stages, potentially increasing log fidelity. Besides that, our work assumes only one activity to appear in an identified video segment at a time. While not a limiting factor for applications like manufacturing lines, this assumption limits the applicability of SnapLog to more complex, diverse scenes where multiple events occur within the same video segment. Finally, at its current stage, SnapLog primarily labels activities and extracts temporal information; we lose spatial information that might be relevant for downstream process analysis and optimization. For example, having spatial information available, one could minimize transportation distances in a manufacturing line in order to increase its efficiency. As of now, such information is not extracted by SnapLog.

Future research will explore using the captured uncertainty for downstream error correction. For example, an LLM could be used as a post-processor to judge the quality of the probabilistic traces and suggest corrections based on its general knowledge of process logic. Additionally, we aim to evaluate whether these video-derived process models can be utilized in related applications, such as video reasoning and predictive machine failure, where high-level activity sequences are vital for operational insights. 
To mitigate the limitations of assuming that only one activity appears in each segment, SnapLog can be extended to not only identify segments relevant to activities throughout the temporal dimension, but also throughout the spatial dimension, ultimately allowing activities to co-occur in a single video segment.

\begin{credits}
\subsubsection{\ackname}
The authors gratefully acknowledge the German Federal Ministry of Education and Research (BMBF) and the Ministry of Education and Research of North-Rhine Westphalia for supporting this work as part of the NHR funding.

\subsubsection{\discintname}
The authors have no competing interests to declare.
\end{credits}
%
%
%
\bibliographystyle{splncs04}
\bibliography{bibliography_dblp}

\end{document}